\documentclass[lettersize,journal]{IEEEtran}
\usepackage{amsmath,amsfonts}
\usepackage{algorithmic}
\usepackage{algorithm}
\usepackage{array}
\usepackage[caption=false,font=normalsize,labelfont=sf,textfont=sf]{subfig}
\usepackage{textcomp}
\usepackage{stfloats}
\usepackage{url}
\usepackage{verbatim}
\usepackage{graphicx}
\usepackage{cite}
\usepackage[pagebackref=false,breaklinks=false,linkcolor=black,anchorcolor=black, citecolor=black,colorlinks,bookmarks=true]{hyperref}
\usepackage{multirow}
\usepackage[table,xcdraw]{xcolor}
\usepackage{booktabs}
\usepackage[normalem]{ulem}
\useunder{\uline}{\ul}{}
\newcommand{\etal}{\textit{et al}. }

\newcommand{\ie}{\textit{i}.\textit{e}., }

\begin{document}

\title{SplatFlow3D: Learning Scene Flow via\\3D Motion Feature Splatting}

\author{
	Bo Wang, 
	Zhenping Sun, 
	Yang Yu, 
	Jian Li,~\IEEEmembership{Member,~IEEE}, \\
	Li Liu,~\IEEEmembership{Senior Member,~IEEE}, 
	and Dewen Hu,~\IEEEmembership{Senior Member,~IEEE}
	\thanks{
		\textit{Corresponding authors: Li Liu and Dewen Hu.}
	}
	\thanks{Bo Wang is with the Defense Innovation Institute, Chinese Academy of Military Science, Beijing, China.}
	\thanks{Zhenping Sun, Yang Yu, Jian Li, and Dewen Hu are with the College of Intelligence Science and Technology, National University of Defense Technology, Changsha, China.}
	\thanks{Jian Li and Dewen Hu are also with the School of Future Technology, National University of Defense Technology, Changsha, China.}
	\thanks{Li Liu is with the College of Electronic Science and Technology, National University of Defense Technology, Changsha, China.}
	\thanks{Email: wb@alumni.nudt.edu.cn, sunzhenping@nudt.edu.cn, yuyangnudt@hotmail.com, lijian@nudt.edu.cn, liuli\_nudt@nudt.edu.cn, dwhu@nudt.edu.cn. }
	\thanks{Copyright © 2026 IEEE. Personal use of this material is permitted. However, permission to use this material for any other purposes must be obtained from the IEEE by sending an email to pubs-permissions@ieee.org. }
}

\markboth{IEEE TRANSACTIONS ON CIRCUITS AND SYSTEMS FOR VIDEO TECHNOLOGY}%
{Wang \MakeLowercase{\textit{et al.}}: SplatFlow3D}

\maketitle

\begin{abstract}
	Considering the motion continuity and stability of 3D objects over short inter-frame intervals, introducing multi-frame settings in scene flow estimation can potentially address occlusion. 
	In this paper, we propose a two-frame backbone, SimpleFlow3D, which integrates dense information from the image plane and sparse information from the depth dimension in a decoupling-followed-by-coupling manner by hierarchically constructing the 3D motion feature. SimpleFlow3D achieves SOTA results with minimal parameters. 
	We also propose a weight-sharing multi-frame framework to splat the previous 3D motion feature to the current estimation process. The framework unifies the first and non-first frame inference by constructing pseudo inputs. 
	After combining the proposed framework and backbone, the newly born multi-frame method SplatFlow3D further advances the state of the art. It achieves lower errors than the two-frame backbone, with fewer parameters than current methods. 
	On both KITTI and Spring, SplatFlow3D ranked 1st over all published and submitted methods. 
	Experiments show that SplatFlow3D improves occlusion handling, non-rigid motion estimation, and generalization to real-world scenes. 
	Code is available at \href{https://github.com/wwsource/SplatFlow3D}{https://github.com/wwsource/SplatFlow3D}. 
\end{abstract}

\begin{IEEEkeywords}
	Multi-frame Settings, Scene Flow Estimation, Occlusion, Non-rigid Motion, Splatting Transformation. 
\end{IEEEkeywords}

\section{Introduction}

Accurately estimating the 3D motion of objects plays a critical role in understanding complex scene structures and behavioral patterns, and therefore finds extensive applications such as human motion estimation \cite{niu2024method, yu2023toward}, object tracking \cite{yang2022temporal, zhong2017direct}, and moving object segmentation \cite{shao2018motion}. 
As a fundamental and challenging problem in 3D computer vision, scene flow estimation (SFE) aims to estimate the pixel-wise/point-wise 3D motion between consecutive stereo, RGB-D, or point cloud data, thereby providing more challenging dense and fine-grained 3D motion estimation results. 

Existing methods employ dense feature point matching: using a deep neural network to extract the higher-order feature representations of pixels \cite{opticalexp, raft3d, rigidmask} or LiDAR points \cite{flownet3d, flot} and then directly matching or indirectly regressing dense 3D motion based on the principle of feature similarity.

Feature similarity is only applicable to non-occluded regions between frames, while occlusion is widespread in real-world scenarios; hence, SFE faces a stern test.
To address this issue, most methods mainly focus on introducing rigid motion priors through explicit \cite{gordon2019depth, rigidmask} or implicit \cite{raft3d} approaches. 
These priors allow occluded regions to enjoy a unified and orderly 3D transformation as part of the entire rigid body. 
However, powerful semantic representation abilities are required, which can lead to an increase in model parameters and significant data requirements. 

Considering the motion continuity and stability of 3D objects over short inter-frame intervals, introducing multi-frame settings can also address occlusion. 
However, this comes with certain risks. The inference of each frame requires processing information from multiple frames, which significantly increases the computational load. In addition, there is a potential conflict between the priors provided by historical frames and the judgment of the current frame, leading to unstable, or even worse, estimation results. 

\textcolor{black}{
In our research, we aim to design a simple yet effective multi-frame scene flow method, SplatFlow3D, which requires the ability to effectively address occlusion and non-rigid motion. To this end, we propose two key components: a two-frame backbone and a multi-frame framework. }

First, we propose a novel iterative two-frame backbone, SimpleFlow3D. 
The network integrates dense information from the image plane and sparse information from the depth dimension in a decoupling-followed-by-coupling manner by hierarchically constructing the 3D motion feature. 
Specifically, we construct the appearance correlation feature based on RGB images and the depth correlation feature based on depth maps, respectively. While characterizing the distribution of correlations, these two features can be used to indirectly regress the plane displacement and directly calculate the depth change. 
We propose a novel convolutional layer that uniformly encodes the two features with optical flow and depth change to generate a 3D motion feature, which can serve as a carrier to simultaneously store the 3D motion estimate and confidence. The entire network iteratively encodes 3D motion features and outputs motion residuals to update predictions. 

Second, we propose a weight-sharing multi-frame framework for the proposed scene flow backbone. 
Inspired by the recent optical flow method SplatFlow \cite{splatflow}, we introduce the splatting transformation, \ie a differentiable forward warping operation, to unidirectionally transmit the previous frame's 3D motion feature to each iteration of the current frame. 
We propose sharing weights for the first and non-first frame inference by constructing pseudo inputs. This significantly reduces the number of network parameters and further improves network performance by accelerating convergence. 
We combine this framework with the proposed backbone to realize a new multi-frame scene flow method, SplatFlow3D. 

In benchmark experiments, the proposed two-frame SimpleFlow3D achieves state-of-the-art (SOTA) results with minimal parameters. 
Furthermore, the proposed multi-frame SplatFlow3D improves on these SOTA results, with fewer parameters than current methods. 
On FlyingThings3D \cite{Things}, compared with the best published RGB-D input-based method, SplatFlow3D reduces the 3D error from 0.064 to 0.036, with a 43.8\% error reduction. 
On KITTI2015 \cite{KITTI} and Spring \cite{Spring}, at the time of submission, SplatFlow3D ranked 1st over all published and submitted methods. 

The study offers the following contributions:
\begin{itemize}
\item We propose a novel two-frame method, SimpleFlow3D, which integrates dense information from the image plane and sparse information from the depth dimension in a decoupling-followed-by-coupling manner by hierarchically constructing the 3D motion feature. 
\item We propose a new multi-frame method, SplatFlow3D, by combining the proposed backbone with a weight-sharing multi-frame framework. The framework unifies the first and non-first frame inference by constructing pseudo inputs. 
\item The two methods achieve SOTA results on all mainstream scene flow datasets with minimal parameters. 
\item Experiments show that SplatFlow3D improves occlusion handling, non-rigid motion estimation, and generalization to real-world scenes. 
\end{itemize}

\section{Related Work}

\subsection{Two-frame Scene Flow Estimation Methods}

Traditional approaches for estimating two-frame scene flow usually leverage the variational technique, which can be solved as an energy minimization process \cite{huguet2007variational, KITTI, wang2015handling}. 
For instance, Wang \etal \cite{wang2015handling} develop an over-parameterized scene flow representation that models both rotational and translational motion components, coupled with a two-stage optimization framework to estimate two-frame scene flow. 
The emergence of deep learning has enabled the direct extraction of feature representations and patterns from data, leading to remarkable progress in scene flow estimation \cite{opticalexp, ISF, ACOSF, DRISF, raft3d, rigidmask, CamLiFlow}. 

Existing methods generally rely on feature similarities. For example, FlowNet3D \cite{flownet3d} operates directly on point clouds with PointNet++ \cite{pointnet++}, with a flow embedding computed in one layer to capture the correlation between two point clouds, which is then propagated through finer layers to estimate the scene flow. 
The cutting-edge CamLiFlow \cite{CamLiFlow} consists of 2D \cite{pwcnet} and 3D \cite{pointpwc} branches with multiple bidirectional connections in specific layers. Both branches adopt a pyramid structure, which refines the optical flow and scene flow from the two frames’ feature similarity tensors in a coarse-to-fine way.
Some variations in the principle of feature similarity exist. For example, OpticalExp \cite{opticalexp} uses optical expansion to recover the pixel’s motion in depth between frames. By combining the initial depth value of the first frame, it can estimate the final depth value of the second frame, indirectly estimating the scene flow per pixel. However, the calculation of optical expansion comes from accurately estimating optical flow \cite{VCN}, which still relies on feature similarity. 

To address occlusion, existing methods introduce rigid motion priors. Most recent work \cite{DRISF, ISF, gordon2019depth, rigidmask} introduces explicit rigid motion priors to the network through additional object detection or segmentation. 
For example, RigidMask \cite{rigidmask} assumes that the scene is made up of a background and several dynamic rigid bodies. These instances are segmented through a dedicated network, combined with rough initial scene flow estimation, and independently parameterized by 3D rigid transformations.
Although it performs well on driving datasets like KITTI \cite{KITTI}, it requires pixel-level ground truth annotation to train another moving rigid body segmentation network. 

Unlike the above methods, RAFT-3D \cite{raft3d} implicitly incorporates rigid motion priors in its network. It uses semantic information to classify each pixel into a soft group of rigid objects and solves the optimized 3D rigid transformation of the group through a differentiable Gauss-Newton algorithm. 
The combination of GRU and Gauss-Newton iterations, together with the large number of pixels involved in the solution of each objective, determines an enormous theoretical computational cost. 
Thanks to customized CUDA operators, it enjoys a high degree of computational optimization.
It is noted that this high degree of customization also implies that the maintenance and improvement of the method may pose some challenges. 

Compared with existing methods, our approach dose not use rigid motion priors, which can be highly constrained by scene-specific information and semantic details. Instead, we have developed a carefully designed network that can more effectively utilize multi-frame information that is more universal in nature. This allows for greater computational efficiency and improved performance across a wider range of scenes and circumstances.

\begin{figure*}[t]
	\begin{center}
		\includegraphics[width=\linewidth]{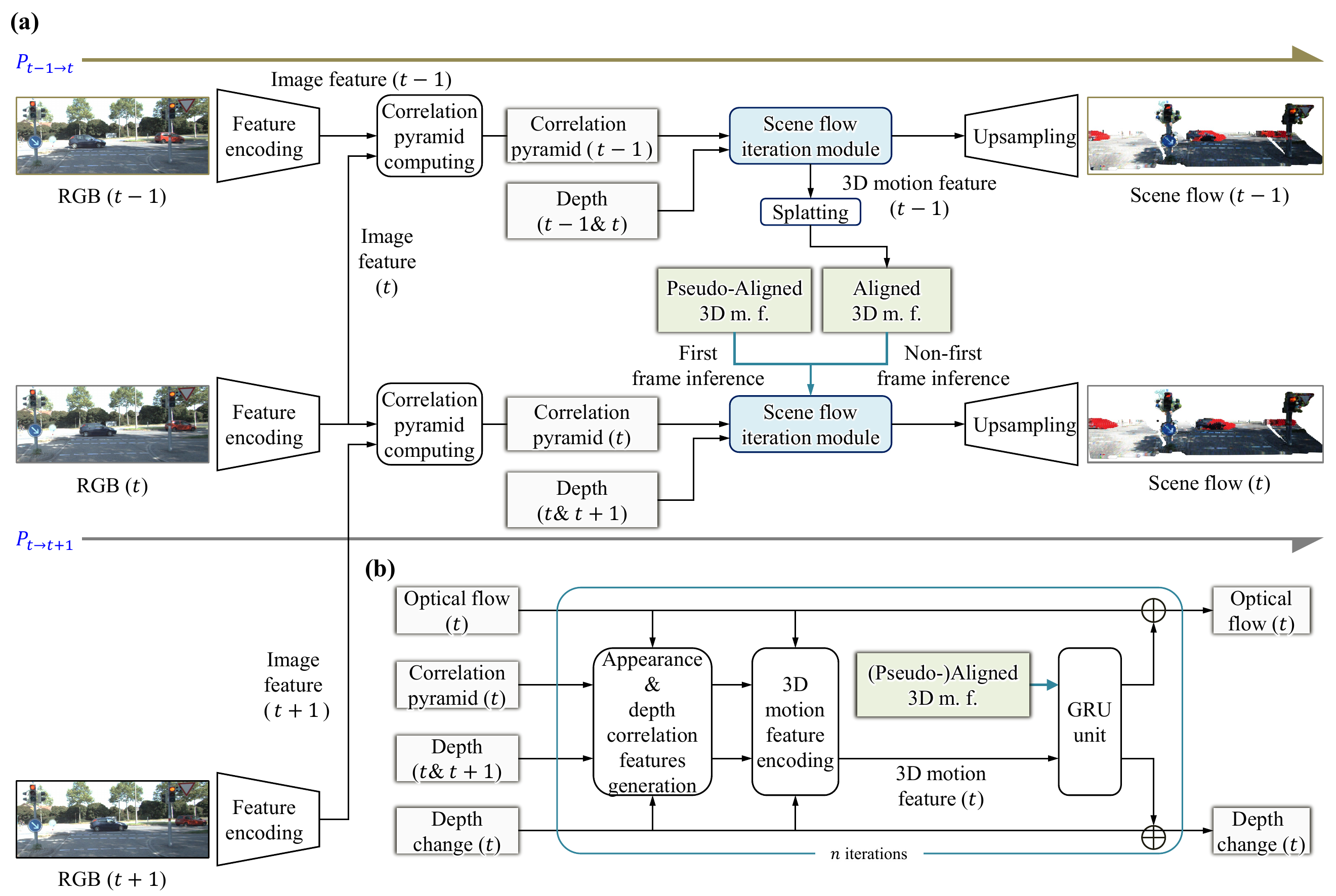}
	\end{center}
	\caption{
		\textbf{Architecture of proposed method.} 
		\textbf{a}, SFE processes from frame $t-1$ to frame $t$ and from frame $t$ to frame $t+1$; \textbf{b}, Scene flow iteration module. 
	}
	\label{fig: overview}
\end{figure*}

\subsection{Multi-frame Scene Flow Estimation Methods}

Multi-frame information can provide richer visual and motion priors, and is widely used in depth estimation \cite{feng2023iterdepth, wu2023self}, optical flow estimation \cite{splatflow}, and scene flow estimation. 
For example, Feng \etal \cite{feng2023iterdepth} propose a context-aware transformer to refine the matching costs generated from multiple frames, and Wu \etal \cite{wu2023self} extract depth features from consecutive frames and compute the 3D cost volume. Multi-frame information is used here to improve the accuracy of depth estimation. 

Traditional multi-frame SFE methods \cite{vogel20153d, golyanik2017multiframe, taniai2017fast, neoral2017object, schuster2020sceneflowfields++} also use the energy minimization framework to explicitly model multi-frame scene flow. 
Built upon the piece-wise rigidity assumption, Vogel \etal \cite{vogel20153d} ensure a temporally consistent piece-wise planar segmentation. 
While following a similar conceptual framework, Golyanik \etal \cite{golyanik2017multiframe} employ RGB-D data as opposed to stereo sequences. 
Taniai \etal \cite{taniai2017fast} propose a fusion framework that integrates estimates derived from optical flow and multi-frame stereo matching across temporal sequences. 
Building upon Menze and Geiger \cite{KITTI}'s two-frame method, Neoral and Šochman \cite{neoral2017object} enhance the approach through the integration of temporal object label propagation. 
Schuster \etal \cite{schuster2020sceneflowfields++} propose a sparse-to-dense approach that accelerates energy-based computation through sparse multi-frame matching, assuming temporal consistency in 3D scene flow motion, followed by robust interpolation. 

Deep neural networks have gained significant traction and demonstrated remarkable success in the domain of multi-frame scene flow estimation. 
DTF \cite{dtf} employs a compact, learnable motion inverter network to predict the inverted forward flow from the backward flow. Subsequently, it integrates the forward flow and the inverted forward flow through a convex fusion step, ultimately yielding the final scene flow estimation. 
Hur and Roth \cite{hur2021self} introduced a self-supervised monocular approach that leverages a convolutional LSTM network to ensure temporal consistency over time. 
M-FUSE \cite{m-fuse} draws inspiration from DTF \cite{dtf} by utilizing a neural network for flow prediction fusion. It is built upon the cutting-edge two-frame backbone network, RAFT-3D \cite{raft3d}, with its architecture meticulously designed to align seamlessly with RAFT-3D’s framework. Notably, the network employs local \textit{SE}(3) parameterization combined with inputs derived from RAFT-3D’s recurrent units to effectively integrate temporal information. Furthermore, it introduces an \textit{SE}(3)-based extrapolation operation and a generalized U-Net-based fusion mechanism, both of which significantly enhance its performance and adaptability across diverse scenarios. 

\section{Proposed Method}

Fig. \ref{fig: overview} shows the architecture of our method. 
The thick brown and black arrows in Fig. \ref{fig: overview}a show the SFE process of the proposed SplatFlow3D from frame $t-1$ to frame $t$ (denoted as $P_{t-1\rightarrow t}$) and from frame $t$ to frame $t+1$ (denoted as $P_{t\rightarrow t+1}$), respectively. 

\subsection{Preliminaries}

Our method follows the recurrent correlation-based pipeline of RAFT \cite{raft}. This pipeline has three standard operations. We use them as they are. We describe them below. 

\textbf{Correlation pyramid construction.} RGB data are encoded into image features by a convolutional network consisting of six residual blocks and three downsampling layers. Given the image features $F_t^i$ and $F_{t+1}^i$ of frames $t$ and $t+1$, the per-pixel correlation volume is computed by the dot product of the two feature maps. The last two dimensions of this volume are average-pooled with kernels of size 1, 2, 4, and 8. This gives a four-layer correlation pyramid $Y_t$. 

\textbf{Correlation sampling.} At the $n$th iteration, the $(n-1)$th optical flow provides a target position for each pixel. Around this position, each layer of $Y_t$ is bilinearly sampled on a grid of size $(2\times r+1)^2$, where $r$ is the local neighborhood radius. The sampled values of all layers are concatenated into the appearance correlation feature. 

\textbf{GRU-based update.} The appearance correlation feature and the current state are fed into a GRU unit. The GRU unit uses convolutional layers to update the hidden state feature, and then regresses a residual. The residual updates the motion fields. In our method, the motion fields are the optical flow and the depth change. 

\subsection{SimpleFlow3D}

The inference process of the proposed two-frame backbone SimpleFlow3D is equivalent to the $P_{t\rightarrow t+1}$ process of removing the previous frame's information input (blue arrow, Fig. \ref{fig: overview}a). 

SimpleFlow3D takes the RGB-D data of frames $t$ and $t+1$ as input. 
The RGB data are encoded into image features by the encoder described in Section III-A. 
The correlation pyramid of frame $t$ is built by the standard construction in Section III-A. 

The correlation pyramid and the depth maps of frames $t$ and $t+1$ are input to a scene flow iteration module to output the iteratively updated optical flow and depth change. 
Then, the network uses convex upsampling to generate the full-resolution results, which can be further converted into the scene flow of frame $t$ with camera intrinsics. 

\begin{figure}[h]
	\begin{center}
		\includegraphics[width=\linewidth]{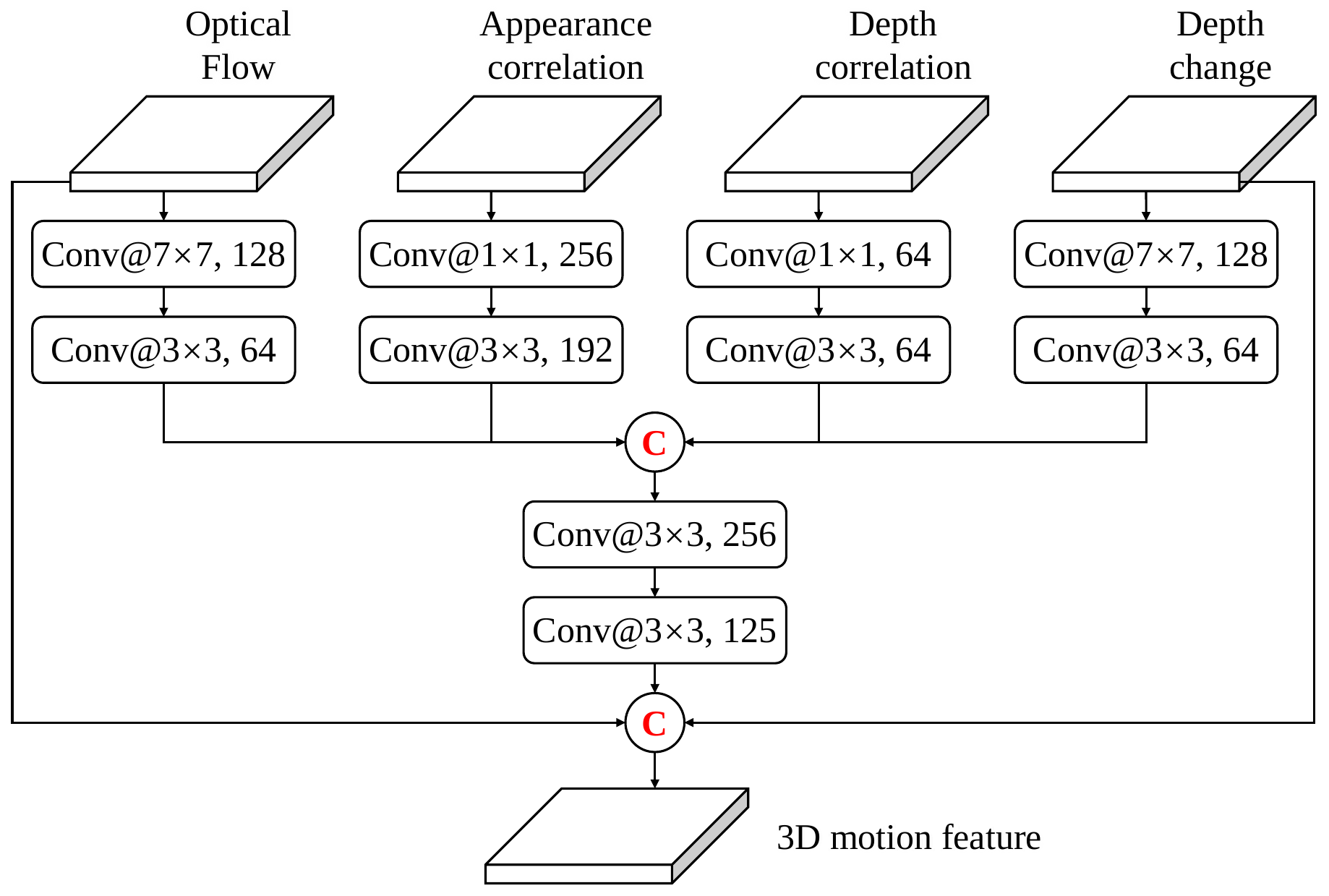}
	\end{center}
	\caption{
		Schematic illustration of the whole encoding process of the 3D motion feature. 
		$\mathrm{Conv}$@$n\times n,l$: convolution operation with kernel $n\times n$ and output channels $l$, followed by a ReLU operation. 
		\textbf{\textcolor{red}{C}}: concatenation operation. 
	}
	\label{fig: mf}
\end{figure}

Data enter the scene flow iteration module (Fig. \ref{fig: overview}b) $n$ times iteratively. 
Taking the $n$th time as an example, the network generates the appearance correlation feature and depth correlation feature, respectively. 
The appearance correlation feature is obtained by the standard sampling in Section III-A. 
The construction process of the depth correlation feature is as follows: 
(1) We use the ($n-1$)th optical flow to perform $(2\times r+1)^2$ grid-based bilinear sampling on the depth map of frame $t+1$, and sequentially concatenate these sampled depth values to obtain the depth sampling map. (2) We sum the ($n-1$)th depth change and the depth map of frame $t$ to generate the depth prediction map. (3) We compute inverse depths for both the depth sampling map and depth prediction map, respectively, then subtract them to obtain the depth correlation feature. 
Subsequently, a convolutional network jointly encodes the optical flow, depth change, and two correlation features to generate a 3D motion feature. 
Fig. \ref{fig: mf} shows the 3D motion feature encoding process. 
With the 3D motion feature as input, the GRU unit performs the standard update in Section III-A. 

The above describes the proposed decoupling-followed-by-coupling integration approach for image plane and depth dimension data, which first constructs the appearance and depth correlation features separately and then encodes them into the 3D motion feature uniformly. 

In our final model, we follow GMA \cite{gma} to globally aggregate the 3D motion feature using self-similarity, to obtain the spatially aggregated 3D motion feature as additional input to the GRU unit. 
For clarity, SimpleFlow3D adds the depth correlation feature and 3D motion feature extractions, unlike RAFT \cite{raft} and GMA \cite{gma}. It iteratively updates the image plane and depth dimension motion fields instead of updating the \textit{SE}(3) motion field, as in RAFT-3D \cite{raft3d}. 

\subsection{SplatFlow3D}

Building upon SimpleFlow3D, we introduce the proposed multi-frame SplatFlow3D, which uses a weight-sharing multi-frame framework to unify the first and non-first frame inference. 

If $t > 1$, the multi-frame framework will establish temporal connections between the estimation process $P_{t-1\rightarrow t}$ of the previous frame and the estimation process $P_{t\rightarrow t+1}$ of the current frame for the proposed two-frame backbone SimpleFlow3D, and use the Splatting transformation to construct the aligned 3D motion feature from the previous frame as additional temporal information input for the current frame $t$. 
If $t = 1$, the multi-frame framework will construct a pseudo-aligned 3D motion feature as additional temporal information input for the current frame $t$.
The network employs shared weights to process the same-sized additional inputs constructed for both scenarios, thus achieving unified inference. \\

\subsubsection{Non-first Frame Inference}

As illustrated in Fig. \ref{fig: overview}a, we employ the Splatting transformation to establish a bridge for transferring historical information from $P_{t-1\rightarrow t}$ to $P_{t\rightarrow t+1}$. 
When performing the non-first frame inference, we splat the 3D motion feature of frame $t-1$ to the coordinate system of frame $t$ to obtain the aligned 3D motion feature, which is then fed into the scene flow iteration module's all iterations, as additional input to the GRU unit. All inputs of the GRU unit will be concatenated into a new input feature for updating the hidden state feature. 

In the process of splatting, the source feature will publish its value to the neighboring integer pixels of its target sub-pixel, with the 2D motion direction provided by the optical flow. The value published to the same integer pixel will be weighted and summed in a specific normalization to generate a complete target feature map. 
Specifically, we use Softmax Splatting as our splatting type, \ie we use an exponential value of the source-to-target similarity to calculate normalized weights in our SplatFlow3D. 
The mathematical expression of the normalized weight is as follows,
\begin{equation}
    n_i=\frac{exp(Z_i) \cdot w_i}{\sum_{j=1}^{k} exp(Z_j) \cdot w_j} \label{eq:ni_softmax},
\end{equation}
where $i$ means the $i^{th}$ value published to the same integer pixel, $w_i$ is the bilinear weight, and $Z_i$ is the source-to-target similarity. 
As a crucial component, the differentiable splatting transformation strongly supports the proposed multi-frame framework, and further efficiently addresses the challenges of occlusion and multi-frame computational burden in scene flow estimation.

\subsubsection{First Frame Inference} 

If frame $t$ is the initial frame of the entire video, the aforementioned operations cannot be directly executed. 
Given the absence of historical frames, it would be na\"ive to use another dedicated two-frame network for the first frame inference, such as the approach of SplatFlow \cite{splatflow}. However, the two independent networks lead to redundancy in the model structure, and they increase learning difficulty. 

Based on the above considerations, the proposed weight-sharing multi-frame framework uses only one GRU unit to handle all scenarios. 
For the first frame inference, we construct a pseudo-aligned 3D motion feature to match the input types and quantities of the non-first frame inference, as shown in Fig. \ref{fig: overview}b. Although the input composition of the GRU unit remains identical in both inference scenarios, the network needs to adaptively distinguish between these two cases. This distinction is crucial because the pseudo-aligned 3D motion feature not only fails to provide valuable motion priors but also may introduce noise interference, thereby affecting the estimation accuracy of both the first frame and subsequent frames. To address this, we employ a zero-initialized tensor with dimensions identical to the aligned 3D motion feature for constructing the pseudo-aligned 3D motion feature. This approach ensures that the pseudo-aligned 3D motion features constructed for the first frame of all videos are identical, which reduces the network's difficulty in differentiating and learning this specific scenario. Through this approach, the network employs exactly the same architecture and weights when performing the first and non-first frame inference. 

\subsection{Training and Inference} \label{section: loss}

The difference between multi-frame settings and two-frame settings lies in that multi-frame settings require input information from more than two frames to estimate the scene flow of the current frame. 
Our method additionally embeds information from the previous frame, \ie we use three frames in multi-frame settings. 
Whenever the next frame of data is acquired, the network calculates the scene flow from the current frame to the next frame, and constructs the aligned 3D motion feature for embedding into the scene flow estimation of the next frame. 

We only need to train on short sequence data composed of just three frames to achieve inference on long sequence data composed of any number of frames. 
Specifically, we utilize the data of three consecutive frames as a training unit, with the optical flows and inverse depth changes of the first two frames serving as supervision for our network. Given the ground truth optical flows $g_{t-1}^{uv}$ and $g_{t}^{uv}$, and ground truth inverse depth changes $g_{t-1}^{id}$ and $g_{t}^{id}$, we define the loss as follows:
\begin{align}
    \mathcal{L}
    &=\beta \cdot \mathcal{L}_{t-1} + \mathcal{L}_{t}\nonumber\\
    &=\beta \cdot \sum_{k=1}^{n} \gamma^{n-k} (||g_{t-1}^{uv}-o_{t-1,k}^{uv}||_1 + \alpha \cdot ||g_{t-1}^{id}-o_{t-1,k}^{id}||_1)\nonumber\\
    &+\sum_{k=1}^{n} \gamma^{n-k} (||g_{t}^{uv}-o_{t,k}^{uv}||_1 + \alpha \cdot ||g_{t}^{id}-o_{t,k}^{id}||_1) \label{eq:Loss1},
\end{align}
Here we set $\gamma=0.85$, $\alpha=1000$, and $\beta=0.75$. 
\textcolor{black}{
Considering that only frame $t$ in the data unit of KITTI \cite{KITTI} contains the ground truth, 
we use this to alternately supervise the two inference processes of embedding the pseudo-aligned and aligned 3D motion feature with $\mathcal{L}=\mathcal{L}_{t}$ as the loss when training KITTI data. }

\section{Experiments}
\subsection{Experimental Setup}

\subsubsection{Datasets}

The primary datasets for scene flow tasks are FlyingThings3D \cite{Things} (hereinafter Things), KITTI2015 \cite{KITTI} (hereinafter KITTI), and Spring \cite{Spring}. Our method requires three adjacent frames to be merged into a single input data unit for network training. Consistent with the approach used in SplatFlow \cite{splatflow}, we adjust Things \cite{Things} and KITTI \cite{KITTI}. 
Regarding Things \cite{Things}, considering only the left camera, we can collect a training dataset containing 71,648 data units. We use the middle three frames of each forward testing sequence to form a multi-frame testing dataset. The clean pass and final pass each consist of 437 data units. 
\textcolor{black}{
For KITTI \cite{KITTI}, we use the multi-view extension of the standard KITTI 2015 scene flow dataset. }

\subsubsection{Implementation Details}

We use PyTorch to implement our method. 
We use the AdamW \cite{loshchilov2017decoupled} as an optimizer and the One-cycle \cite{smith2019super} as the learning rate adjustment strategy. We evaluate our method after 12 (for training) and 24 (for testing) GRU iterations. We use the NVIDIA RTX 3090 GPU for training and testing. 

\subsubsection{Training Processes}

The training processes of our SimpleFlow3D and SplatFlow3D both include C+T and K-finetune. 
In the C+T process, we first load and freeze pretraining weights and separately train unique sub-modules. Then, we release all frozen parameters and fine-tune all of them. The batch size is 8, there are 100K training steps, and the learning rate is $1.25 \times 10^{-4}$. 
The pretraining weights of SimpleFlow3D come from GMA, and those of SplatFlow3D come from SimpleFlow3D. In addition, we can also train all weights of SplatFlow3D from scratch with 400k training steps and a batch size of 8, to achieve the same performance. 
In the K-finetune process, we fine-tune the model obtained from the C+T process on a KITTI data-enhanced hybrid dataset composed of Things \cite{Things} and KITTI \cite{KITTI}. 
The batch size is 6, the learning rate is $1.25 \times 10^{-4}$, and there are 50K training steps. Considering that the background in KITTI is a rigid body, we follow CamLiFlow \cite{CamLiFlow} and refine the background scene flow using a rigidity refinement step. 

\begin{table*}[t]
	\centering
	\caption{
		Comparison results with current SOTA methods on the KITTI \cite{KITTI} benchmark after the K-finetune training process. 
		Best results of all methods are in boldface, and second-best results are underlined. 
	}
	\label{tab: kitti-test}
	\resizebox{0.75\textwidth}{!}{%
		\begin{tabular}{@{}cccccccccc@{}}
			\toprule
			& \multicolumn{2}{c}{Disparity 1} & \multicolumn{2}{c}{Disparity 2} & \multicolumn{2}{c}{Optical Flow} & \multicolumn{2}{c}{Scene Flow} &  \\ \cmidrule(lr){2-9}
			\multirow{-2}{*}{Method} & Noc & All & Noc & All & Noc & All & Noc & All & \multirow{-2}{*}{Parameters} \\ \midrule
			OpticalExp \cite{opticalexp} & {\ul 1.63} & {\ul 1.81} & 2.62 & 4.25 & 3.89 & 6.30 & 5.21 & 8.12 & 18.5M \\
			DRISF \cite{DRISF} & 2.35 & 2.55 & 3.14 & 4.04 & 3.58 & 4.73 & 4.99 & 6.31 & 58.9M \\
			RAFT-3D \cite{raft3d} & {\ul 1.63} & {\ul 1.81} & 2.67 & 3.67 & 3.23 & 4.29 & 4.53 & 5.77 & 51.3M \\
			RigidMask \cite{rigidmask} & 1.70 & 1.89 & 2.47 & 3.23 & 2.54 & 3.50 & 3.73 & 4.89 & 145.3M \\
			M-FUSE \cite{m-fuse} & \textbf{1.51} & \textbf{1.65} & 2.31 & 3.13 & 2.48 & 3.46 & 3.61 & 4.83 & 49.0M \\
			RAFT-MSF++ \cite{sun2026raft} & - & {\ul 1.81} & - & 2.97 & - & 3.55 & - & 4.75 & 53.4M \\
			CamLiFlow \cite{CamLiFlow} & {\ul 1.63} & {\ul 1.81} & 2.37 & 2.95 & 2.40 & 3.10 & 3.55 & 4.43 & 19.7M \\
			OAMaskFlow \cite{peng2025oamaskflow} & - & {\ul 1.81} & - & 2.89 & - & 3.05 & - & 4.21 & 19.7M \\
			SimpleFlow3D-GANet (Ours) & {\ul 1.63} & {\ul 1.81} & 2.25 & 2.81 & 2.22 & 2.96 & 3.25 & 4.19 & {\ul 19.1M} \\
			SplatFlow3D-GANet (Ours) & {\ul 1.63} & {\ul 1.81} & {\ul 2.24} & {\ul 2.68} & \textbf{2.10} & \textbf{2.87} & {\ul 3.13} & {\ul 4.09} & 19.6M \\
			SplatFlow3D-LEAStereo (Ours) & \textbf{1.51} & \textbf{1.65} & \textbf{2.13} & \textbf{2.61} & {\ul 2.14} & {\ul 2.89} & \textbf{3.12} & \textbf{4.04} & \textbf{15.1M} \\ \bottomrule
		\end{tabular}%
	}
\end{table*}

\subsection{Comparisons with State-of-the-art Methods}

\subsubsection{Results on Things} \label{sec:things}

TABLE \ref{tab: flownet3d} compares our methods with several RGB-D input-based methods on the Things \cite{Things} test dataset using the split provided by FlowNet3D \cite{flownet3d}. 

The results show that our two-frame backbone, SimpleFlow3D, surpasses existing methods on all dataset metrics, which demonstrates the effectiveness of the proposed depth correlation feature construction and 3D motion feature encoding. 
Meanwhile, our multi-frame method SplatFlow3D further achieves better performance, especially in the 43.8\% reduction of $\text{EPE}_{3D}$, from 0.064 for RAFT-3D to 0.036. 
In the above experiments, our methods exhibit superior performance on Things \cite{Things} which features purely rigid scenes. 

\begin{table}[h]
	\centering
	\caption{Scene flow results on Things \cite{Things} using split of FlowNet3D \cite{flownet3d}, where only non-occluded points with depth $<35m$ are considered for evaluation. }
	\label{tab: flownet3d}
	\resizebox{0.45\textwidth}{!}{%
		\begin{tabular}{@{}lllll@{}}
			\toprule
			Method & Input & \multicolumn{1}{c}{$\delta_{3D}^{0.05} \uparrow$} & \multicolumn{1}{c}{$\delta_{3D}^{0.10} \uparrow$} & \multicolumn{1}{c}{$\text{EPE}_{3D} \downarrow$} \\ \midrule
			FlowNet3D++ \cite{flownet3d++} & RGB-D & 30.3 & 63.4 & 0.137 \\
			RAFT-3D \cite{raft3d} & RGB-D & 83.7 & 89.2 & 0.064 \\
			SimpleFlow3D (Ours) & RGB-D & {\ul 86.8} & {\ul 92.3} & {\ul 0.046} \\
			SplatFlow3D (Ours) & RGB-D & \textbf{88.0} & \textbf{93.3} & \textbf{0.036} \\ \bottomrule
		\end{tabular}%
	}
\end{table}

\subsubsection{Results on KITTI} \label{section: kitti}

\begin{table*}[t]
	\centering
	\caption{
		Comparison results with current SOTA methods on the Spring \cite{Spring} benchmark after the C+T training process. 
	}
	\label{tab: spring}
	\resizebox{0.9\textwidth}{!}{%
		\begin{tabular}{cccccccccc}
			\toprule
			Method & $\text{1px}^{SF}$ & $\text{1px}^{matched}$ & $\text{1px}^{unmat.}$ & $\text{1px}^{rigid}$ & $\text{1px}^{non-rig.}$ & $\text{All}^{SF}$ & $\text{1px}^{D1}$ & $\text{1px}^{D2}$ & $\text{1px}^{Fl}$ \\ \midrule
			RAFT-3D \cite{raft3d} & 78.8 & 78.3 & 85.2 & 79.6 & 72.8 & 66.9 & 23.2 & 73.4 & 48.1 \\
			CamLiFlow \cite{CamLiFlow} & 50.1 & 47.8 & 79.6 & 46.8 & 75.3 & 34.2 & 23.2 & 44.1 & 24.0 \\
			M-FUSE \cite{m-fuse} & 34.9 & 32.0 & 71.9 & 29.8 & 73.4 & 16.1 & 19.9 & 24.3 & 20.4 \\
			PWOC-3D \cite{saxena2019pwoc} & 26.7 & 23.4 & 69.3 & 21.8 & 64.0 & 10.0 & {\ul 18.2} & 19.0 & 14.5 \\
			SimpleFlow3D-no-refinement (Ours) & 24.8 & 21.6 & 65.8 & 20.7 & 55.7 & 11.4 & 19.9 & {\ul 15.0} & 9.2 \\
			SimpleFlow3D (Ours) & {\ul 19.8} & {\ul 16.7} & {\ul 59.4} & {\ul 16.7} & {\ul 43.5} & {\ul 7.9} & 19.9 & \textbf{15.0} & {\ul 9.2} \\
			SplatFlow3D (Ours) & \textbf{18.9} & \textbf{15.9} & \textbf{57.1} & \textbf{15.8} & \textbf{42.0} & \textbf{7.9} & \textbf{15.1} & 15.1 & \textbf{7.6} \\ \bottomrule
		\end{tabular}%
	}
\end{table*}

TABLE \ref{tab: kitti-test} compares the results of our methods and current SOTA methods on the KITTI \cite{KITTI} benchmark after the K-finetune training process. 
As seen from TABLE \ref{tab: kitti-test}, when using GANet \cite{ganet} to provide depth estimation, our two-frame backbone SimpleFlow3D surpasses all existing submitted and published methods in terms of SF-all (4.19), with a minimum parameter cost of 19.1M. 
Furthermore, when using LEAStereo \cite{leastereo} to provide depth estimation, our final multi-frame version, SplatFlow3D, has the optimal SF-all (4.04). 
At the same time, the parameters of our final model are also lower than those of existing methods. 

\begin{figure}[t]
	\begin{center}
		\includegraphics[width=\linewidth]{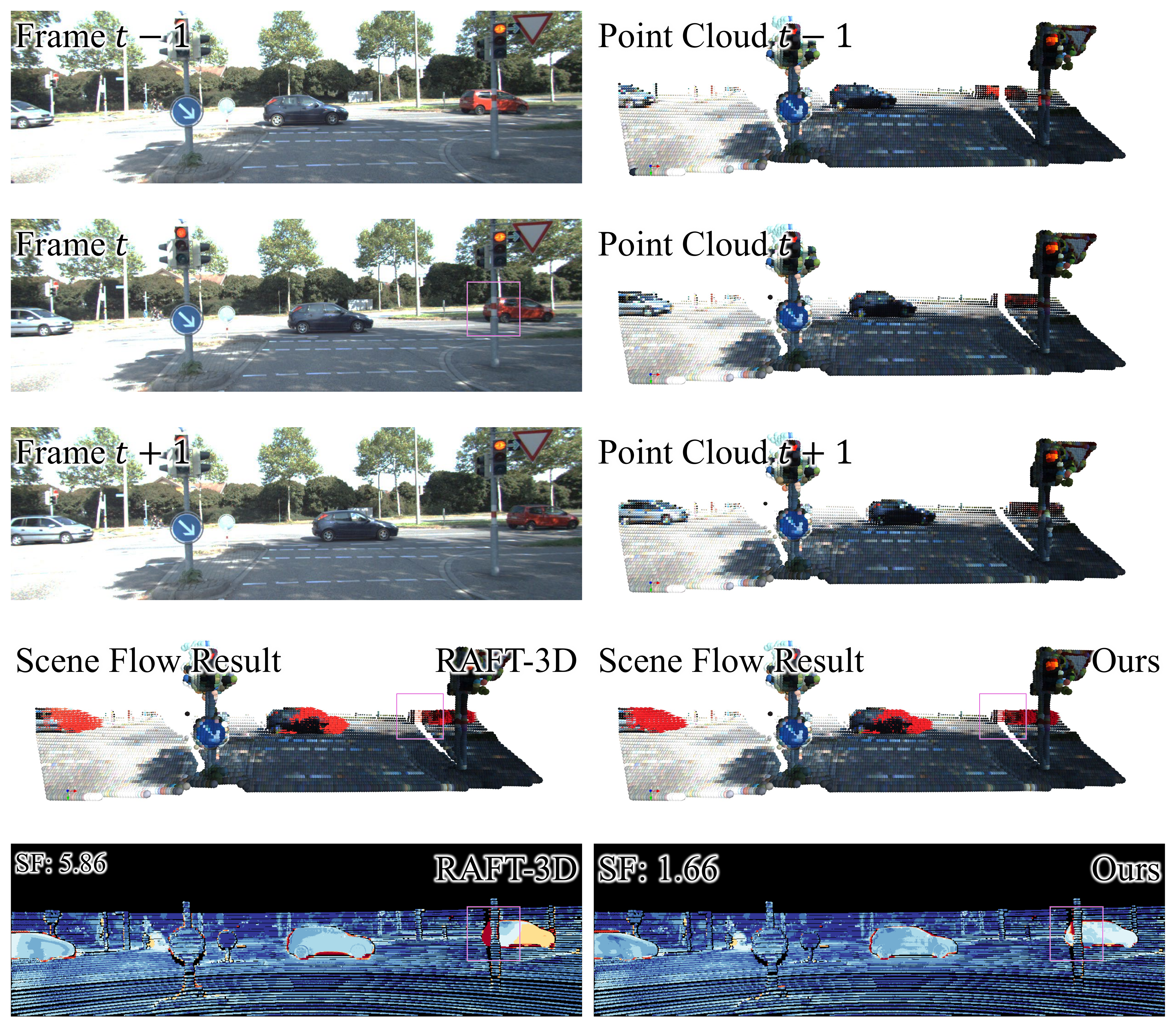}
	\end{center}
	\caption{
		Qualitative results of our method and RAFT-3D \cite{raft3d} 
		on the KITTI \cite{KITTI} test dataset. 
		The region marked by the box has a large appearance difference in the frame $t+1$. 
	}
	\label{fig: kitti}
\end{figure}

Fig. \ref{fig: kitti} shows the qualitative results of SplatFlow3D and RAFT-3D \cite{raft3d} on the KITTI \cite{KITTI} test dataset. The region marked by the box has a large appearance difference in frame $t+1$, which is difficult to estimate. This shows that our method can obtain more stable and reliable estimation results in these challenging regions. 

\subsubsection{Results on Spring}

TABLE \ref{tab: spring} compares the results of our method and current SOTA methods on the Spring \cite{Spring} benchmark after the C+T training process. We use LEAStereo \cite{leastereo} (with weights on Things \cite{Things}) to provide depth estimation, the same as M-FUSE \cite{m-fuse}. Furthermore, we refine the depths by limiting the disparities of LEAStereo to positive numbers.  

As seen from TABLE \ref{tab: spring}, SplatFlow3D achieves the best estimation results on most dataset metrics, especially $\text{1px}^{SF}$ and $\text{1px}^{Fl}$, which are improved by 29.4\% and 47.6\%, respectively, compared with the current best method, PWOC-3D \cite{saxena2019pwoc}. 
Also, we test SimpleFlow3D-no-refinement, which outperforms all existing methods on most dataset metrics even without depth refinements. 
The results show that our methods have significant generalization abilities. 

In addition, compared with PWOC-3D, SplatFlow3D reduces $\text{1px}^{rigid}$ and $\text{1px}^{non-rig.}$ by 27.5\% and 34.5\%, respectively, which shows the advantage of our method on non-rigid motion. 

\subsection{Ablation Studies}

\begin{table*}[t]
	\centering
	\caption{Quantitative results of ablation experiments. All data come from the C+T training process. We evaluate on full image (excluding pixels with depth $>250m$ or flow $>250px$).}
	\label{tab: ablation}
	\resizebox{\textwidth}{!}{%
		\begin{tabular}{@{}cccccccccc@{}}
			\toprule
			\multirow{2}{*}{Settings} & \multirow{2}{*}{Variations} & \multirow{2}{*}{Method} & \multicolumn{3}{c}{Things \cite{Things} Clean} & \multicolumn{3}{c}{Things \cite{Things} Final} & \multirow{2}{*}{\begin{tabular}[c]{@{}c@{}}Paramet-\\ ers (M) $\downarrow$\end{tabular}} \\ \cmidrule(lr){4-9}
			&  &  & $\text{EPE}_{2D} \downarrow$ & $\text{EPE}_{3D} \downarrow$ & $\delta_{3D}^{0.10} \uparrow$ & $\text{EPE}_{2D} \downarrow$ & $\text{EPE}_{3D} \downarrow$ & $\delta_{3D}^{0.10} \uparrow$ &  \\ \midrule
			- & Baseline & GMA \cite{gma} & 4.55 & - & - & 4.01 & - & - & 5.8 \\ \midrule
			\multirow{5}{*}{Two-frame} & \multirow{3}{*}{\begin{tabular}[c]{@{}c@{}}Depth\\ processing\end{tabular}} & Decoupling & 4.55 & 5.545 & 76.5 & 4.01 & 5.568 & 76.6 & \textbf{5.8} \\
			&  & Coupling & 6.85 & 0.228 & 57.6 & 6.81 & 0.228 & 57.5 & 6.3 \\
			&  & Decoupling+Coupling & \textbf{4.38} & \textbf{0.100} & \textbf{85.5} & \textbf{4.00} & \textbf{0.095} & \textbf{85.5} & 7.1 \\ \cmidrule(l){2-10} 
			& \multirow{2}{*}{\begin{tabular}[c]{@{}c@{}}$C^d$ feature\\ channel\end{tabular}} & Single & 4.38 & 0.100 & 85.5 & 4.00 & 0.095 & 85.5 & \textbf{7.1} \\
			&  & Multiple (SimpleFlow3D) & \textbf{4.14} & \textbf{0.087} & \textbf{86.8} & \textbf{4.00} & \textbf{0.085} & \textbf{86.7} & 7.1 \\ \midrule
			\multirow{4}{*}{Multi-frame} & \multirow{2}{*}{\begin{tabular}[c]{@{}c@{}}Temporal\\ Connection\end{tabular}} & w/o & 4.14 & 0.087 & 86.8 & 4.00 & 0.085 & 86.7 & \textbf{7.1} \\
			&  & w & \textbf{2.78} & \textbf{0.067} & \textbf{88.8} & \textbf{2.57} & \textbf{0.065} & \textbf{88.8} & 10.6 \\ \cmidrule(l){2-10} 
			& \multirow{2}{*}{\begin{tabular}[c]{@{}c@{}}Multi-frame\\ framework\end{tabular}} & Independent & 2.78 & 0.067 & 88.8 & 2.57 & 0.065 & 88.8 & 10.6 \\
			&  & Weight-sharing (SplatFlow3D) & \textbf{2.67} & \textbf{0.066} & \textbf{88.8} & \textbf{2.43} & \textbf{0.064} & \textbf{89.0} & \textbf{7.6} \\ \bottomrule
		\end{tabular}%
	}
\end{table*}

We conduct a series of ablation experiments. 
We start from GMA \cite{gma} as the baseline, which iteratively regresses the residual flows of the image plane, and gradually replace its components with our proposed components. All data come from the C+T training process. 

\subsubsection{Depth Processing}

Optical flow estimation methods \cite{zhai2019optical, raft, wang2024acr} only need to process dense data from 2D planes. 
The sparse nature of the depth dimension determines that scene flow estimation is not a straightforward extension of optical flow estimation from 2D to 3D. 
We design two straightforward depth processing methods from a 2D perspective. Among them, ``Decoupling'' utilizes the optical flow of GMA to index the depth map of the next frame, thereby calculating the depth change. Inspired by GMA, ``Coupling'' encodes RGB-D data uniformly and iteratively regresses the residual flows of the $XYZ$ dimensions. 

Unlike the two baselines, we first generate the appearance correlation feature and the single-channel depth correlation feature, which together serve as the 3D motion confidence. We then encode them, along with the 3D motion estimate, uniformly into the 3D motion feature in a decoupling-followed-by-coupling manner. We refer to this baseline as ``Decoupling+Coupling''. 

It can be seen from TABLE \ref{tab: ablation} that ``Decoupling'' exhibits significantly higher 3D error in $\text{EPE}_{3D}$. This is because directly using optical flow for indexing is only effective for non-occluded regions. 
However, after directly copying GMA, ``Coupling'' still does not achieve satisfactory results on the 2D and 3D dataset metrics. This indicates that SFE differs from OFE, and it cannot simply apply OFE techniques to handle the additional depth dimension. 
Finally, we can see that ``Decoupling+Coupling'' shows significant advantages over the other two baselines across all 2D and 3D dataset metrics, which fully demonstrates the effectiveness of the proposed method. 
Furthermore, combined with Section \ref{sec: temporal connection}, we can derive an important insight, that is, a feature, which stores both 3D motion estimate and confidence, can significantly improve the regression accuracy of scene flow in both two-frame and multi-frame settings. 

\subsubsection{Depth Correlation Feature Channel}

In SimpleFlow3D, we construct the depth correlation feature (denoted as $C^d$) solely based on the depth data before encoding the unified 3D motion feature. Based on the different sampling radii $r^d$ of $C^d$, we design the ``Single'' and ``Multiple'' baselines with radii of 0 and 4, respectively. 

It can be seen from TABLE \ref{tab: ablation} that ``Multiple'', \ie SimpleFlow3D, is better than ``Single'' on all dataset metrics, at a similar parameter cost. 
The performance improvement indicates that large-scale sampling can enhance the positioning accuracy of both the image plane and depth dimension. 

\subsubsection{Temporal Connection} \label{sec: temporal connection}

When extending the proposed two-frame backbone, SimpleFlow3D, to multi-frame settings, we design the baseline that splats the previous frame's 3D motion feature to the current frame's estimation process with two independent networks for different situations' inference. 

From TABLE \ref{tab: ablation}, we find that after adding the temporal connection, the baseline achieves significant performance improvements on all six dataset metrics, especially $\text{EPE}_{3D}$, which reduces the average error of 23.3\%. 
This result demonstrates the effectiveness of multi-frame information in improving the estimation effect of scene flow. 

\subsubsection{Multi-frame Framework}

We rename the multi-frame baseline proposed in Section \ref{sec: temporal connection} to ``Independent'' because it uses another independent two-frame network for the first frame inference. 
We propose a new weight-sharing multi-frame framework using a single network for both inferences. 

From TABLE \ref{tab: ablation}, we find that ``weight-sharing'', \ie the proposed SplatFlow3D, outperforms ``Independent'' over all metrics, especially parameter cost, which is reduced by 28.3\%. These results highlight that the proposed multi-frame framework can better unify the first and non-first frame inference, demonstrating the functional redundancy of the two independent networks and the advantages of constructing pseudo inputs. 

\begin{figure}[h]
	\begin{center}
		\includegraphics[width=\linewidth]{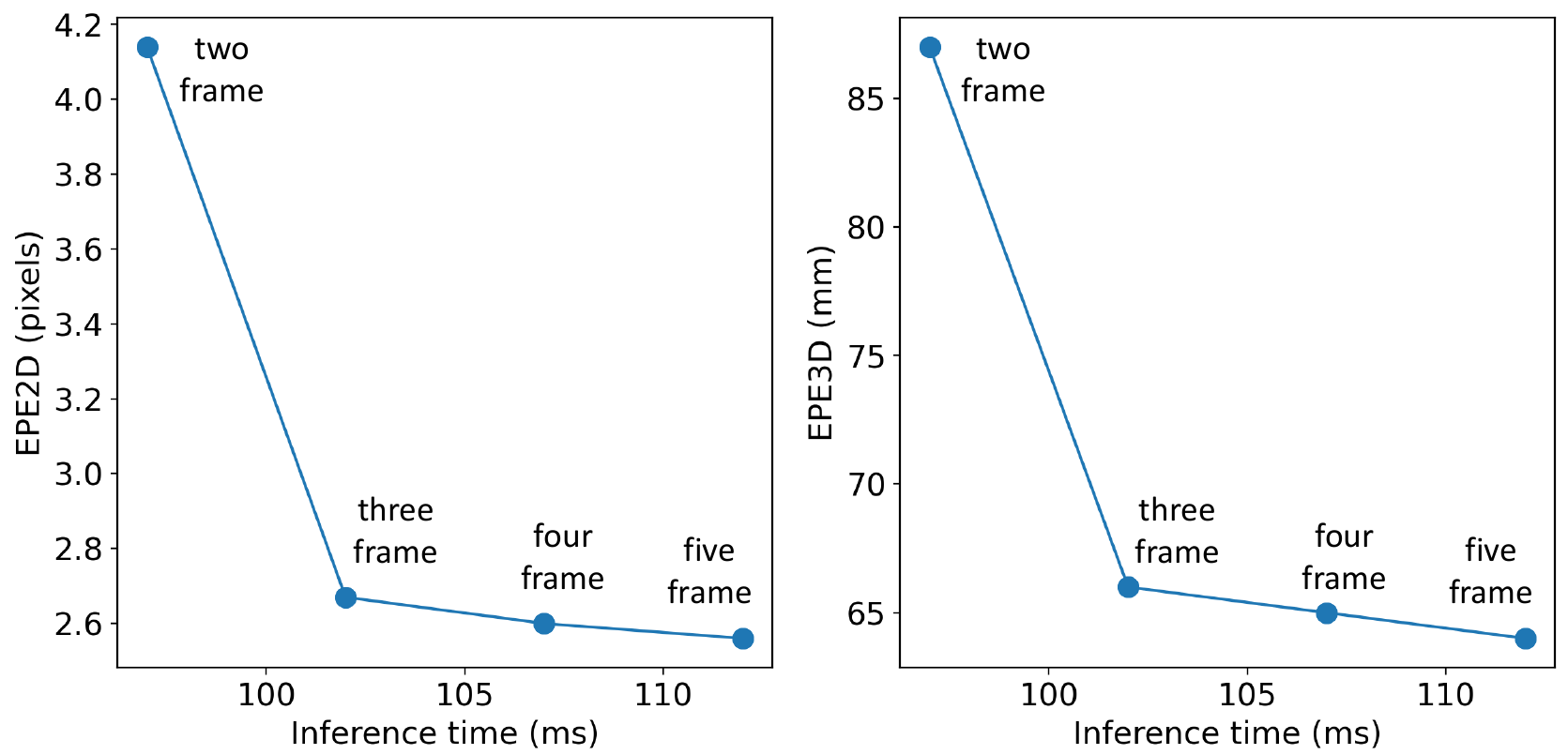}
	\end{center}
	\caption{
		Evaluation results on the Things val dataset versus the inference time as the number of frames increases. 
	}
	\label{fig: frame_num}
\end{figure}

\subsubsection{Number of Frames}

We extend our SplatFlow3D with more frames to explore the effect of increasing the number of frames on the performance of scene flow estimation. 
We use optical flows $\{O_{t-k\rightarrow t-k+1}, O_{t-k+1\rightarrow t-k+2}, ..., O_{t-1\rightarrow t}\}$ to sequentially splat the 3D motion feature of frame $t-k$ to frame $t$, thereby obtaining the aligned 3D motion feature $A^{3D}_{t-k\rightarrow t}$. Next, the feature set $\{A^{3D}_{t-k\rightarrow t}, A^{3D}_{t-k+1\rightarrow t}, ..., A^{3D}_{t-1\rightarrow t}\}$ is concatenated into a new input feature for the GRU unit. 

We evaluate the models on the Things val dataset after the C+T training process. The results are shown in Fig. \ref{fig: frame_num}. 
As shown in Fig. \ref{fig: frame_num}, our multi-frame versions all significantly outperform the original two-frame version as the number of frames increases. 
This fully demonstrates the high scalability of the proposed multi-frame framework and indicates that introducing multi-frame settings ($\geq$3) is beneficial for improving the performance of scene flow estimation. 
Moreover, the introduction of a third frame significantly reduces 2D and 3D estimation errors by 35.5\% and 24.1\%, respectively, while the performance gains from further increasing the number of frames diminish sharply ($\textless$3\% per frame). The three-frame configuration therefore achieves the optimal balance between precision and efficiency. 

\subsubsection{Comparisons with Multi-frame Baselines}

\begin{table}[h]
	\centering
	\caption{
		\textcolor{black}{
			Evaluation results and relative improvements of multi-frame methods compared with two-frame backbones on the KITTI \cite{KITTI} benchmark. 
			We keep the same depth source as M-FUSE, \ie backbone uses GANet, and multi-frame version uses LEAStereo. 
			We do not refine the background scene flow. 
		}
	}
	\label{tab: kitti-multi-frame}
	\resizebox{0.48\textwidth}{!}{%
		\begin{tabular}{@{}crrr@{}}
			\toprule
			Method & \multicolumn{1}{c}{SF Before} & \multicolumn{1}{c}{SF After} & \multicolumn{1}{c}{Rel. Impr.} \\ \midrule
			DTF \cite{dtf} +PWOC \cite{saxena2019pwoc} & 15.69 & 14.64 & 6.7\% \\
			DTF \cite{dtf} +SENSE \cite{jiang2019sense} & 9.55 & 9.18 & 3.9\% \\
			DTF \cite{dtf} +SimpleFlow3D & \textbf{5.52} & 5.66 & -2.5\% \\
			M-FUSE \cite{m-fuse} & {\ul 5.77} & {\ul 4.83} & \textbf{16.3\%} \\
			SplatFlow3D (Ours) & \textbf{5.52} & \textbf{4.71} & {\ul 14.7\%} \\ \bottomrule
		\end{tabular}%
	}
\end{table}

In this subsection, we compare the proposed method with other multi-frame methods to provide a fundamental study exploring where our strong results come from. 

We use DTF \cite{dtf} as a backbone-independent baseline to extend three backbones, including our SimpleFlow3D, and use M-FUSE \cite{m-fuse} as a backbone-dependent baseline, which exploits the unique $SE(3)$ parameterization of RAFT-3D \cite{raft3d}. 
DTF and M-FUSE are both originally multi-frame methods. 
TABLE \ref{tab: kitti-multi-frame} shows the evaluation results, from which it is seen that multi-frame settings can promote SFE. 
In addition, our multi-frame framework can achieve competitive performance and does not require computing a backward scene flow at high time cost like M-FUSE, showing its advantages in mining multi-frame information and utilizing the characteristics of the backbone network. 

Finally, by combining the multi-frame framework with the SOTA backbone, we improve on the two-frame results on multiple benchmarks. 

\subsubsection{Results about 3D Motion Confidence}

We further check whether the 3D motion feature carries the motion confidence of a point.
We evaluate our model on the Things test Clean set. 
For every point we store two $9\times9$ windows at the last GRU iteration. 
The first is the correlation window sampled around the current flow estimate, and we take the peak of the softmax over its 81 values. 
The second is the inverse-depth residual window between the two frames, and we take its absolute value at the position where that peak lies.
This residual is measured in 1/m. 
The two windows have a clear meaning. 
When the match of a point is unique, the correlation window has one sharp peak and the inverse-depth residual window has one clear minimum. 
When the match is ambiguous, both windows are flat.  

\begin{table}[ht]
	\centering
	\caption{
		Relation between the two reliability values and occlusion. The rule flags a point as unreliable, and its threshold maximizes Youden's J over all points. 
	}
	\label{tab:occ_corr}
	\setlength{\tabcolsep}{2pt}
	\begin{tabular}{cccccl}
		\toprule
		Value & \begin{tabular}[c]{@{}c@{}}Occluded\\median\end{tabular} &
		\begin{tabular}[c]{@{}c@{}}Non-occluded\\median\end{tabular} &
		\begin{tabular}[c]{@{}c@{}}Within-sample\\AUC\end{tabular} &
		\begin{tabular}[c]{@{}c@{}}Flag\\threshold\end{tabular} \\ \midrule \midrule
		\begin{tabular}[c]{@{}c@{}}correlation\\peak\end{tabular} & 0.067 & 0.497 & 0.865 & $< 0.198$ \\
		\begin{tabular}[c]{@{}c@{}}inverse-depth residual\\at the peak\end{tabular} & 0.1197 & 0.0001 & \textbf{0.918} & $> 0.0048$ \\ \bottomrule
	\end{tabular}
\end{table}

It can be seen from TABLE \ref{tab:occ_corr} that occluded points have a low correlation peak. 
The median is 0.067 for occluded points and 0.497 for non-occluded points. 
We find the same trend for this residual. 
The median is 0.1197 for occluded points and 0.0001 for non-occluded points. 
The within-sample AUC is 0.865 for the correlation peak and 0.918 for the inverse-depth residual. 
The AUC of a value is the probability that a random occluded point and a random non-occluded point are ranked in the right order by that value.  
A point flagged by the inverse-depth residual rule has an $\text{EPE}_{3D}$ of 0.164 m, against 0.026 m for the other points. 
The ratio is 6.42. 
This shows that the 3D motion feature carries the motion confidence of a point. 
It does not merely describe the motion estimate. 

\subsection{Results on Occlusions and Non-rigid Bodies}

\begin{table}[h]
	\centering
	\caption{
		Evaluation results and relative improvements of SplatFlow3D and SimpleFlow3D on three types of regions on the Things \cite{Things} test Clean dataset after the C+T training process. 
	}
	\label{tab: occ}
	\resizebox{0.45\textwidth}{!}{%
		\begin{tabular}{@{}ccccc@{}}
			\toprule
			\begin{tabular}[c]{@{}c@{}}Things \cite{Things}\\ test Clean\end{tabular} & \begin{tabular}[c]{@{}c@{}}Region\\ Type\end{tabular} & \begin{tabular}[c]{@{}c@{}}Simple-\\ Flow3D\end{tabular} & SplatFlow3D & \begin{tabular}[c]{@{}c@{}}Rel. \\ Impr. (\%)\end{tabular} \\ \midrule \midrule
			\multirow{3}{*}{$\text{EPE}_{2D}$} & Noc & 1.70 & \textbf{1.29} & 24.1 \\
			& Occ & 12.17 & \textbf{7.67} & \textbf{37.0} \\
			& All & 4.14 & \textbf{2.67} & 35.5 \\ \midrule
			\multirow{3}{*}{$\text{EPE}_{3D}$} & Noc & 0.046 & \textbf{0.040} & 13.0 \\
			& Occ & 0.278 & \textbf{0.194} & \textbf{30.2} \\
			& All & 0.087 & \textbf{0.066} & 24.1 \\ \bottomrule
		\end{tabular}%
	}
\end{table}

\begin{figure}[h]
	\begin{center}
		\includegraphics[width=\linewidth]{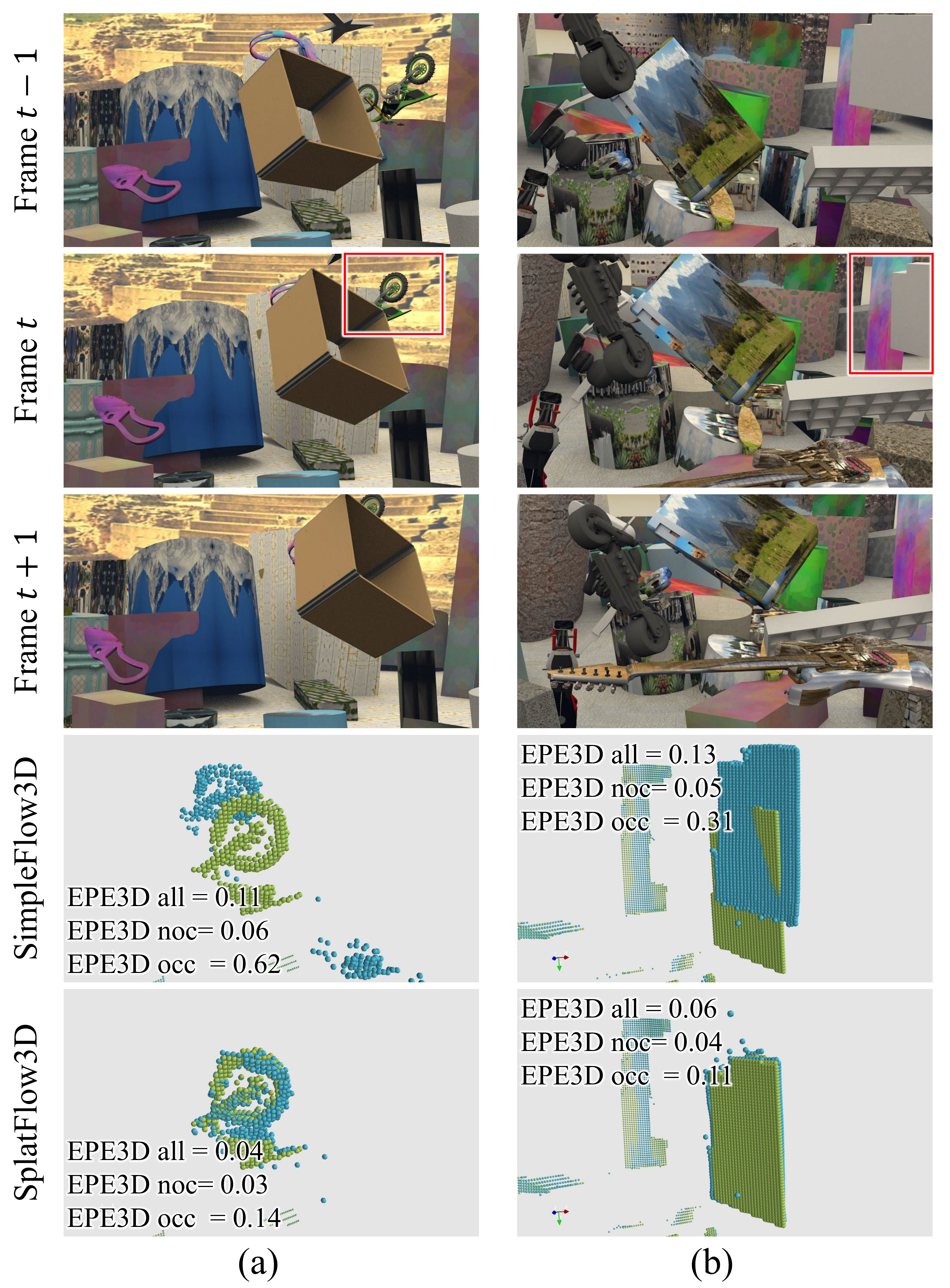}
	\end{center}
	\caption{
		Qualitative results of our methods on the Things \cite{Things} test dataset. 
		Red box: occluded region. 
		Predicted point cloud of frame $t+1$ driven by scene flow is green (ground truth) and blue (two methods). 
	}
	\label{fig: things}
\end{figure}

We analyze the ability of our method to deal with occlusion problems. 
TABLE \ref{tab: occ} shows the evaluation results and relative improvements of SplatFlow3D and SimpleFlow3D on three types of regions (Non-occluded, Occluded, and All) on the Things \cite{Things} test Clean dataset after the C+T training process. 
SplatFlow3D is superior to SimpleFlow3D on all three regions' 2D and 3D dataset metrics. 
Compared with SimpleFlow3D, SplatFlow3D has the most significant improvement in occluded regions. 
Hence, SplatFlow3D can use multi-frame information to comprehensively improve optical and scene flow estimation accuracy, especially in occluded regions. 

Fig. \ref{fig: things} shows the qualitative results of SplatFlow3D and SimpleFlow3D on the Things \cite{Things} test dataset. 
Because these challenging regions are both from the foreground, we only evaluate regions with depth $<35m$. It can be seen that when SimpleFlow3D faces occlusion, its estimation results show obvious deviation, tear, and even fragmentation, while SplatFlow3D can always obtain stable and smooth estimation results, which is also reflected in the dataset metrics of the three regions shown in the subgraphs of Fig. \ref{fig: things}. 

\begin{figure*}[t]
	\begin{center}
		\includegraphics[width=\linewidth]{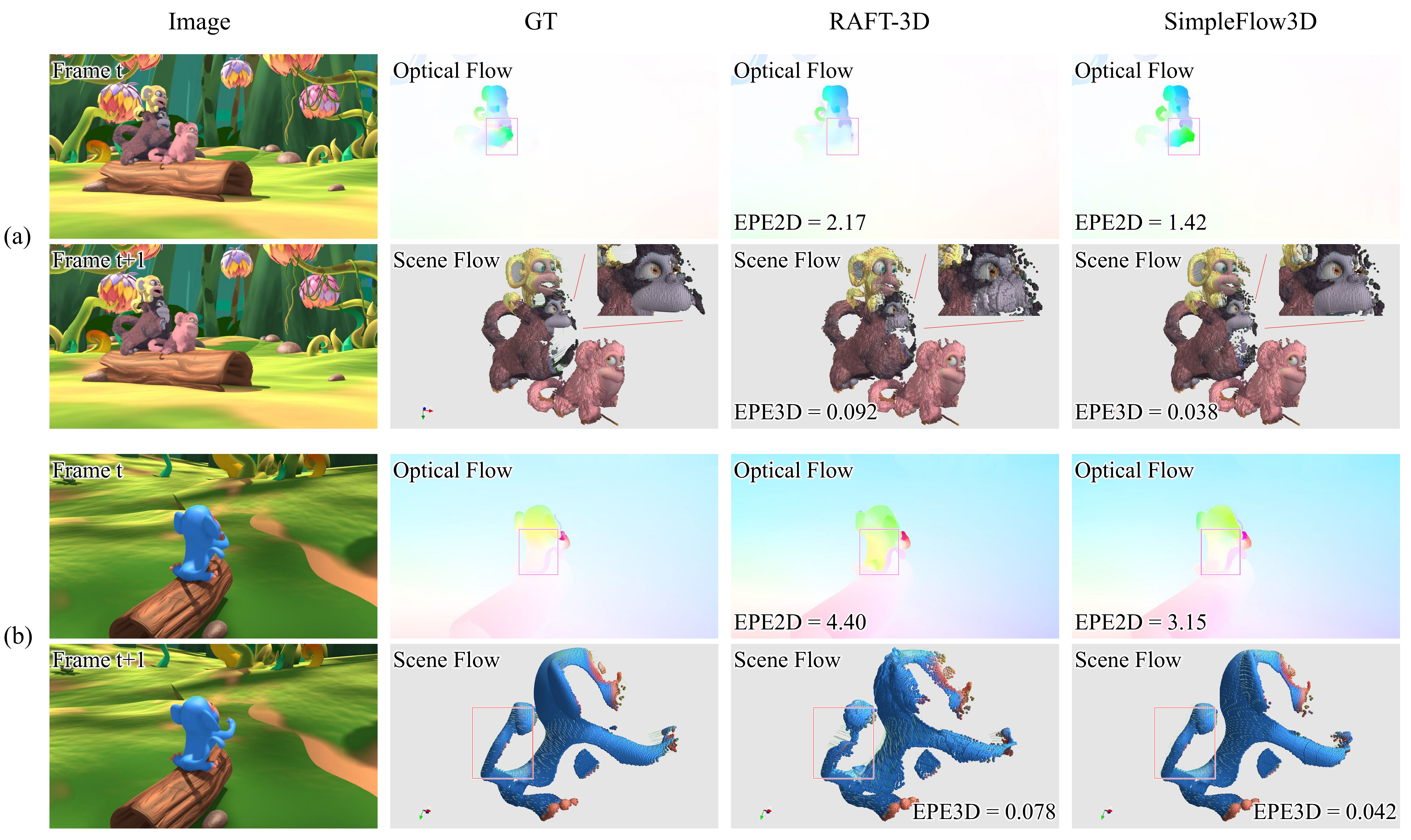}
	\end{center}
	\caption{
		Qualitative results of our SimpleFlow3D and RAFT-3D \cite{raft3d} on the Monkaa \cite{Things} dataset. The region marked by a purple box in the optical flow diagram has independent and different motions from other parts of its instance. The region marked by a red box or lines has significant differences between the predicted results of the two methods. 
	}
	\label{fig: monkaa}
\end{figure*}

Beyond occlusion, non-rigid motion is another key challenge for scene flow estimation. We analyze the ability of our method to deal with it. 
Fig. \ref{fig: monkaa} shows the qualitative results of our SimpleFlow3D and RAFT-3D \cite{raft3d} on Monkaa \cite{Things}, a scene flow dataset containing non-rigid, softly articulated motion and visually challenging fur. 
It can be seen from the optical flow results that the prediction of RAFT-3D \cite{raft3d} in the purple box fails, while our method can output a more accurate estimate. 
We also visualize the scene flow results. We use the image colors of frame $t$ to display the predicted point clouds of frame $t+1$ driven by the scene flows of ground truth and the two methods. 
From the scene flow results, it can be seen that the predicted point cloud of RAFT-3D \cite{raft3d} has a wide range of obvious fluctuations, distortion, and fragmentation, while our method can produce a continuous, smooth, and complete point cloud similar to the ground truth. 

The contrast in outcomes between the two methods arises because RAFT-3D \cite{raft3d} must divide the soft rigid body group according to visual similarity, hampering the motion estimation of each pixel. Our method independently estimates the motion of each pixel.
 
These experiments show that our method also works on non-rigid scenes. 
Moreover, our method can effectively handle unseen non-rigid scenes by solely training on rigid scenes, further substantiating its robustness and strong cross-domain generalization. 

\subsection{Real-world Generalization and Limitations}

\begin{figure}[h]
	\begin{center}
		\includegraphics[width=\linewidth]{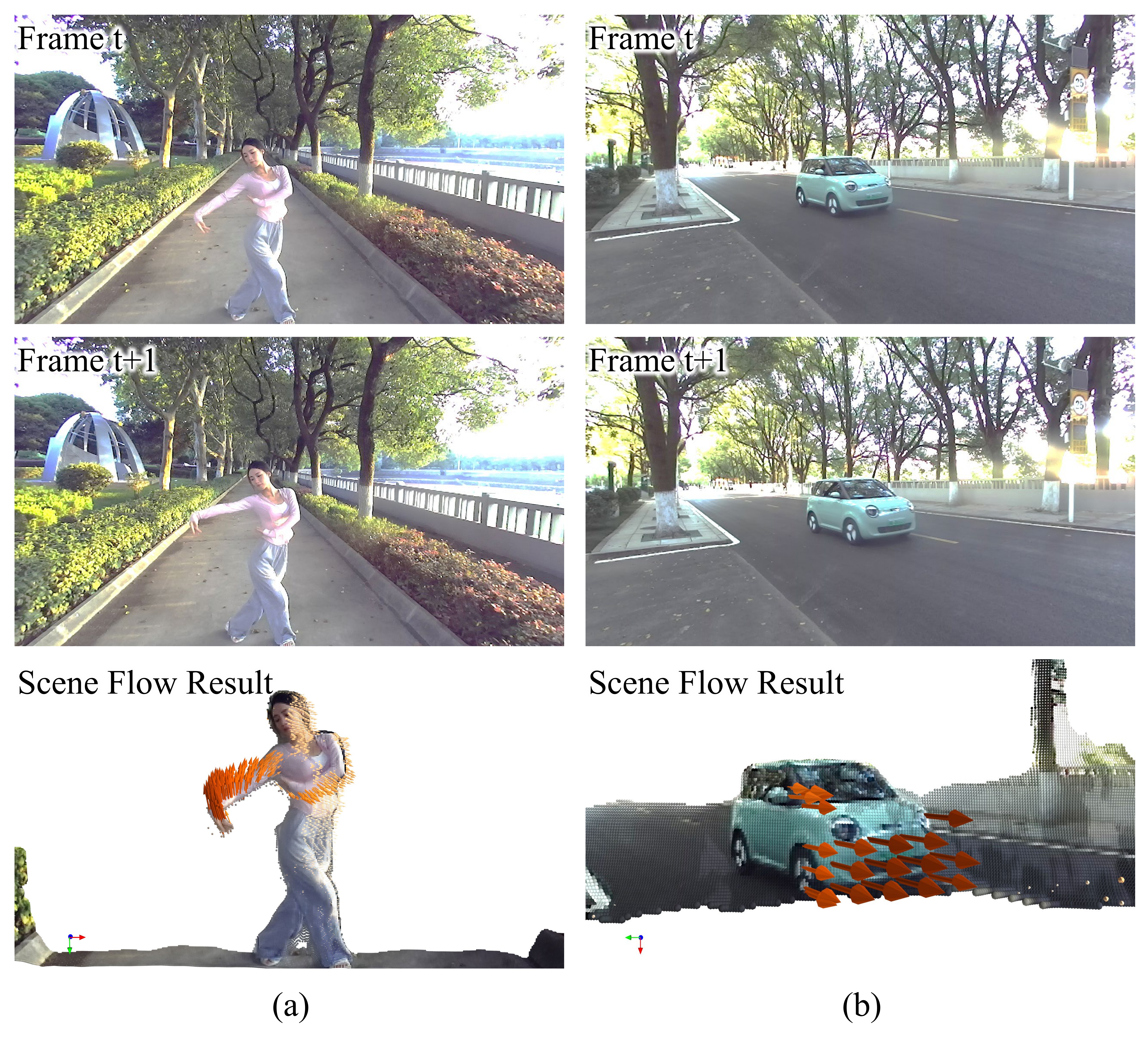}
	\end{center}
	\caption{
		\textcolor{black}{
			Qualitative results of our method on unseen real-world scenes. }
	}
	\label{fig: realworld}
\end{figure}

To further validate the generalization performance of our method, we conduct zero-shot testing on our own collected real-world scene data. We use a ZED binocular camera with 720 $\times$ 1280 resolution and a 0.12-meter baseline to collect data from two real-world scenes. The first video shows a dancer (the left view image is shown in Fig. \ref{fig: realworld}a), and the second video shows a car on a campus road (the left view image is shown in Fig. \ref{fig: realworld}b). 
We use LEAStereo \cite{leastereo} (with weights on KITTI \cite{KITTI}) to provide disparity estimation. 

We display the estimated scene flow results of SplatFlow3D (with weights after K-finetune) in Fig. \ref{fig: realworld}. From the results, it can be seen that our method can effectively estimate both rigid and non-rigid motion in unseen real-world scenes, especially the high-quality capture of complex dance movements, demonstrating the zero-shot generalization performance and non-rigid motion estimation ability of our method. 

Beyond the network design, the results above also rely on the training data. 
In the K-finetune training process, we employ a hybrid dataset combining KITTI data and Things data to train a model for evaluation on real-world datasets. This model achieves the top rank on the KITTI benchmark with a scene flow outlier rate of 4.04. When the model is solely fine-tuned on KITTI data, it only attains an outlier rate of 4.34. These results highlight the significant performance enhancement enabled by large-scale and diverse training data. 

\begin{figure}[h]
	\begin{center}
		\includegraphics[width=\linewidth]{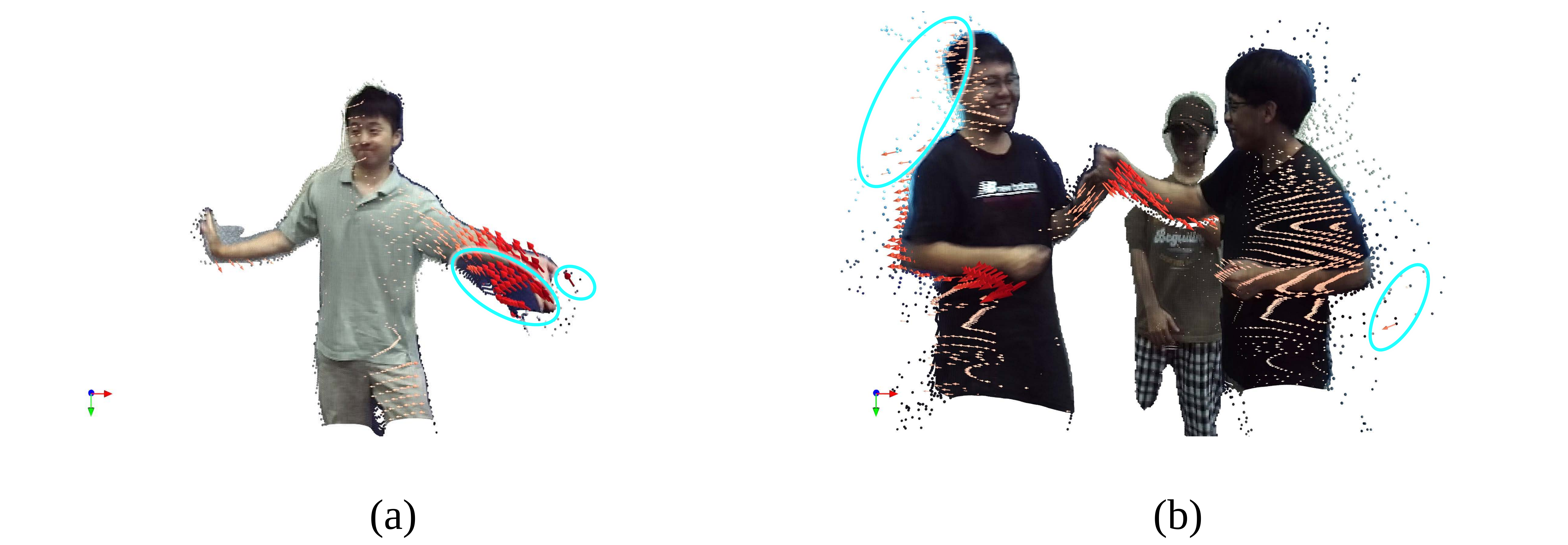}
	\end{center}
	\caption{
		Qualitative results of real-world scenarios where SplatFlow3D exhibits limitations. The region marked by an ellipse has incorrect scene flow estimation results. 
	}
	\label{fig: limitation}
\end{figure}

Furthermore, we visualize real-world scenarios where SplatFlow3D exhibits limitations in Fig. \ref{fig: limitation}. From Fig. \ref{fig: limitation}, it can be seen that most erroneous estimates occur at boundaries between the foreground and background, where binocular depth estimation results are often inaccurate. The aforementioned issue can be mitigated by utilizing more accurate depth estimation results as inputs or adding training samples with depth noise. 

One more limitation comes from the design principle of our method. Our method cannot guarantee that points on the same rigid body have consistent rigid motion. 
This can be improved by introducing semantic information appropriately in the future. 

\section{Conclusion}

In summary, we propose a novel multi-frame scene flow estimation network named SplatFlow3D, 
which splats the previous 3D motion feature to the current estimation process. 
Our method achieves SOTA accuracy, remarkable occlusion handling, non-rigid motion estimation, and real-world scene generalization abilities while maintaining high efficiency in the parameter cost. 

Through this work, we further prove that compared to the scene constraints and semantic module dependencies brought about by the mainstream approach of introducing rigid motion priors, using multi-frame settings has the potential for effective and more general assistance in scene flow estimation. 

\section{Acknowledgement}
This work was partially supported by the National Natural Science Foundation of China under Grants T2521006, U25B2069, 62376283, and 62531026, the Defense Industrial Technology Development Program (JCKY2020550B003) and the Key Stone grant (JS2023-03) of the National University of Defense Technology (NUDT). 

\footnotesize


\newpage

\vspace{-1.0 cm}
\begin{IEEEbiography}[{\includegraphics[width=1in,height=1.25in,clip,keepaspectratio]{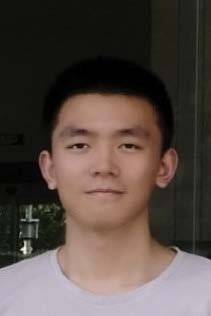}}]{Bo Wang} received the B.E. degree and the Ph.D. degree in control science and engineering from National University of Defense Technology, Changsha, China, in 2019 and 2025. He is currently an assistant researcher with the Defense Innovation Institute, Chinese Academy of Military Science, Beijing, China. His research interests include computer vision and robotics. 
\end{IEEEbiography}

\vspace{-1.0 cm}
\begin{IEEEbiography}[{\includegraphics[width=1in,height=1.25in,clip,keepaspectratio]{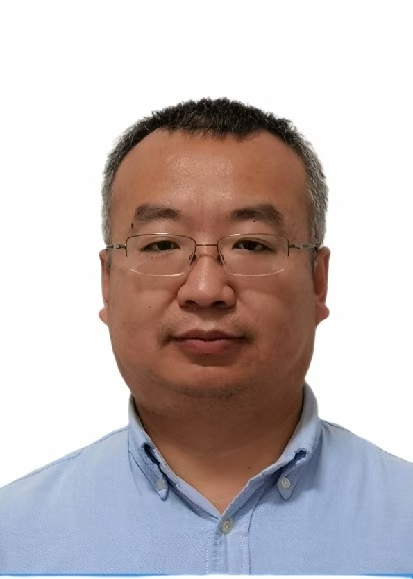}}]{Zhenping Sun} received the B.E. degree and the Ph.D degree in control science and engineering from National University of Defense Technology (NUDT), Changsha, China. He had been a visiting scholar at Karlsruhe Institute of Technology, Germany. He is currently a research fellow of NUDT. His research interests include path planning, motion control, and human-machine interaction of Field robotics. 
\end{IEEEbiography}

\vspace{-1.0 cm}
\begin{IEEEbiography}[{\includegraphics[width=1in,height=1.25in,clip,keepaspectratio]{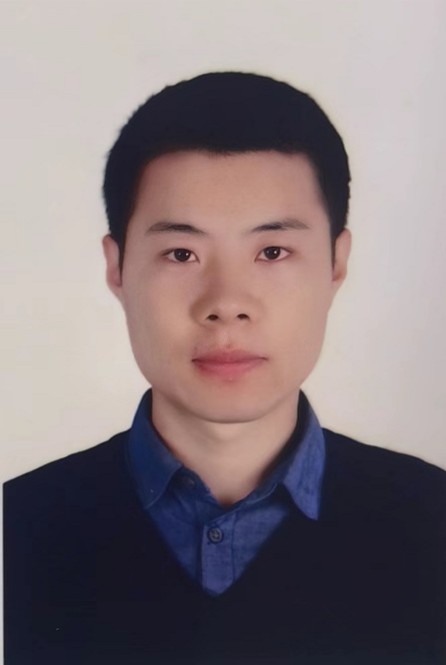}}]{Yu Yang} received the B.S. degree from Tianjin University, Tianjin, China, in 2010. He received the M.S. and Ph.D. degrees from National University of Defense Technology, Changsha, China, in 2012 and 2017, respectively. He is currently an associate professor at the National University of Defense Technology. His research interests include brain–computer interface, machine learning, and neuroimaging.
\end{IEEEbiography}


\vspace{-1.0 cm}
\begin{IEEEbiography}[{\includegraphics[width=1in,height=1.25in,clip,keepaspectratio]{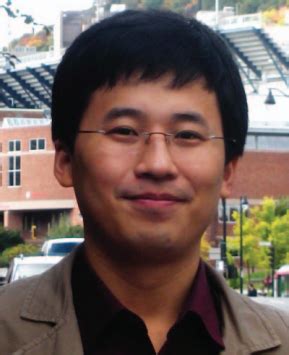}}]{Jian Li} (Member, IEEE) received the B.E. and Ph.D. degrees in control science and engineering from the National University of Defense Technology, Changsha, China. He has been a Visiting Ph.D. Student with the Center for Intelligent Machines, McGill University, Montreal, QC, Canada,
	and a Visiting Scientist for cooperation research with
	the University of Hamburg, Hamburg, Germany. He is currently a Professor with the College of Intelligence Science and Technology, NUDT. His research interests include computer vision, pattern recognition, image processing, and machine learning. He was the recipient of first class Natural Science Award from Hunan Province, China, in 2018. 
\end{IEEEbiography}

\vspace{-1.0 cm}
\begin{IEEEbiography}[{\includegraphics[width=1in,height=1.25in,clip,keepaspectratio]{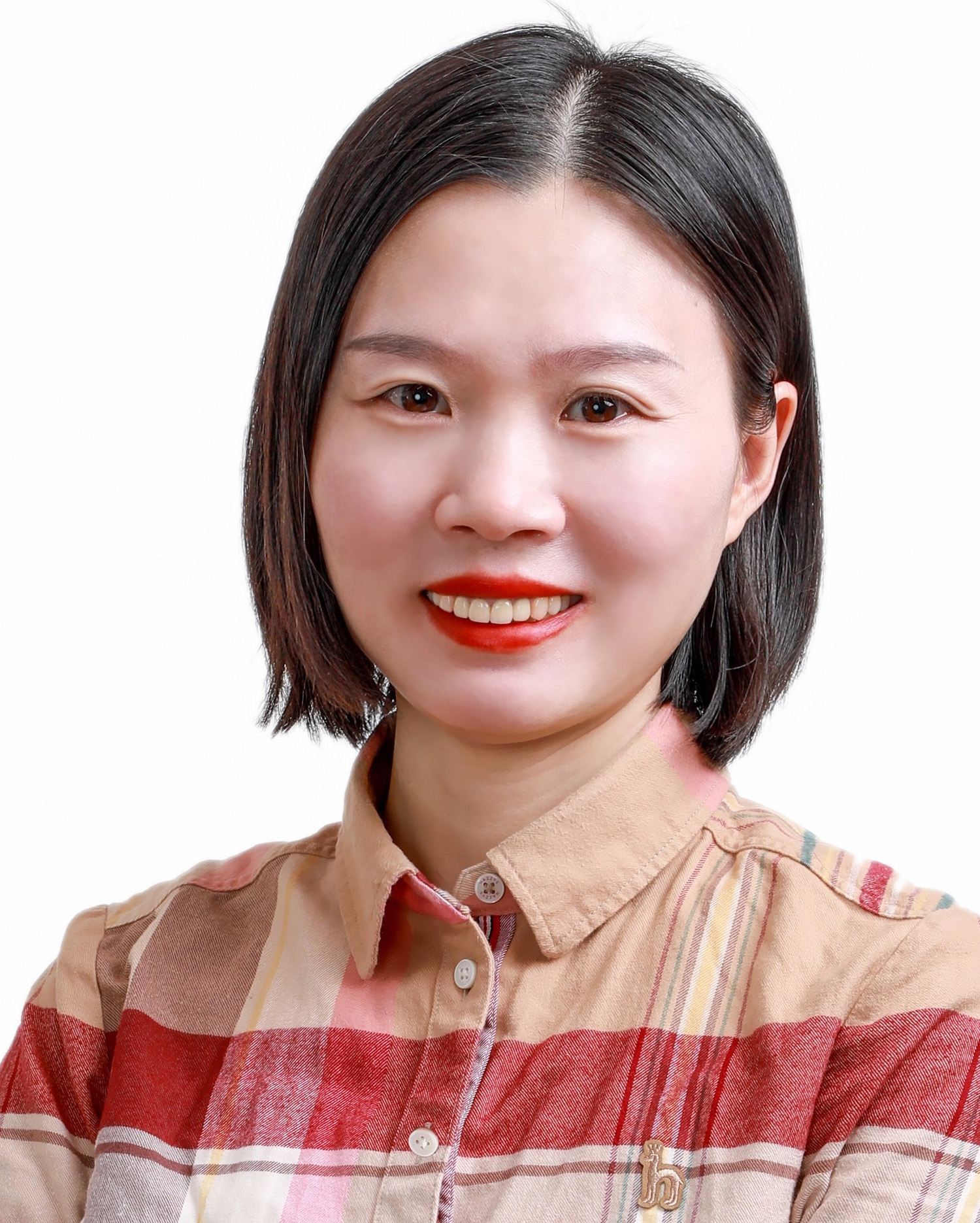}}]{Li Liu} (Senior Member, IEEE) received her Ph.D. degree from the National University of Defense Technology, China, in 2012. She joined the faculty at the National University of Defense Technology in 2012. Dr. Liu was co-chair of nine International Workshops at several major venues, including CVPR, ICCV, and ECCV; she served as the leading guest editor of the special issues for IEEE TPAMI and IJCV. She currently serves as Associate Editor for IEEE TCSVT and IEEE TGRS. Her current research interests include computer vision, pattern recognition, and machine learning. 
\end{IEEEbiography}

\vspace{-1.0 cm}
\begin{IEEEbiography}[{\includegraphics[width=1in,height=1.25in,clip,keepaspectratio]{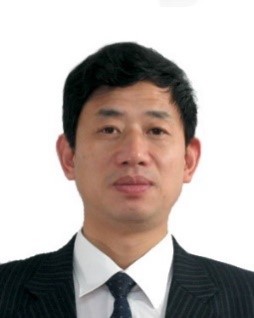}}]{Dewen Hu} (Senior Member, IEEE) received the B.S. and M.S. degrees from Xi’an Jiaotong University, Xi’an, China, in 1983 and 1986, respectively, and the Ph.D. degree from the National University of Defense Technology, Changsha, China, in 1999. He is currently a Professor with the College of Intelligence Science and Technology, National University of Defense Technology. From 1995 to 1996, he was a Visiting Scholar with the University of Sheffield, Sheffield, U.K. He has been an Academician of the Chinese Academy of Sciences since 2025. His research interests include image processing, system identification and control, neural networks, and cognitive science. 
\end{IEEEbiography}

\end{document}